\documentclass{svproc}
\usepackage{amsmath,amssymb,amsfonts}
\usepackage{algorithmic}
\usepackage{graphicx}
\usepackage{textcomp}
\usepackage{xcolor}
\usepackage{booktabs}
\usepackage{pifont}
\usepackage{orcidlink}
\usepackage{hyperref}
\usepackage[ruled,linesnumbered, vlined]{algorithm2e}
\usepackage{fancyhdr}
\usepackage{float}
\newcommand{\cmark}{\ding{51}}%
\newcommand{\xmark}{\ding{55}}%
\usepackage{url}

\begin{document}
\mainmatter              
\title{SSIVD-Net: A Novel Salient Super Image Classification \& Detection Technique for Weaponized Violence}
\titlerunning{SSIVD-Net}  
%
\author{Toluwani Aremu\inst{1} \and Li Zhiyuan\inst{1} \and Reem Alameeri\inst{1} \and Mustaqeem Khan\inst{1} \and Abdulmotaleb El Saddik\inst{1}\inst{2}}
\authorrunning{Aremu, Li, Alameeri, Khan, and El Saddik} 
%
\tocauthor{}
\institute{Mohamed Bin Zayed University of Artificial Intelligence, UAE\\
\and
University of Ottawa, Canada\\
    \email{\{toluwani.aremu, li.zhiyuan, reem.alameeri, mustaqeem.khan, a.elsaddik\}@mbzuai.ac.ae, elsaddik@uottawa.ca}\\
}

\maketitle              

\begin{abstract}
\noindent Detection of violence and weaponized violence in closed-circuit television (CCTV) footage requires a comprehensive approach. In this work, we introduce the \emph{Smart-City CCTV Violence Detection (SCVD)} dataset, specifically designed to facilitate the learning of weapon distribution in surveillance videos. To tackle the complexities of analyzing 3D surveillance video for violence recognition tasks, we propose a novel technique called \emph{SSIVD-Net} (\textbf{S}alient-\textbf{S}uper-\textbf{I}mage for \textbf{V}iolence \textbf{D}etection). Our method reduces 3D video data complexity, dimensionality, and information loss while improving inference, performance, and explainability through salient-super-Image representations. Considering the scalability and sustainability requirements of futuristic smart cities, the authors introduce the \emph{Salient-Classifier}, a novel architecture combining a kernelized approach with a residual learning strategy. We evaluate variations of SSIVD-Net and Salient Classifier on our SCVD dataset and benchmark against state-of-the-art (SOTA) models commonly employed in violence detection. Our approach exhibits significant improvements in detecting both weaponized and non-weaponized violence instances. By advancing the SOTA in violence detection, our work offers a practical and scalable solution suitable for real-world applications. The proposed methodology not only addresses the challenges of violence detection in CCTV footage but also contributes to the understanding of weapon distribution in smart surveillance. Ultimately, our research findings should enable smarter and more secure cities, as well as enhance public safety measures.
\end{abstract}
\section{Introduction}
\label{sec:intro}
\noindent Violence and gang-related activities can pose a serious threat to a city, particularly when authorities are unable to respond quickly enough to prevent further damage. In some cases, these incidents can result in loss of life and property, especially when weapons are involved. Regrettably, incidents of road rage, gang-related violence, and other spontaneous acts of violent crime frequently occur without prior warning or the ability for authorities to intervene proactively. These events pose a considerable challenge for law enforcement agencies and other relevant authorities. Unfortunately, the reporting of such incidents often occurs after the fact, leaving authorities with limited options for timely intervention and effective prevention. 

Although surveillance systems have helped authorities identify instigators and culprits through recordings, it often takes too long to detect, search, and arrest someone after a crime is committed. To reduce turnaround time and increase efficiency, there is a growing need for automated detection and signaling systems. Since the breakthrough of deep learning \cite{2.alexnet} in the ImageNet 2012 competition, deep neural networks (DNNs) have become the go-to AI technology for automating such tasks. By leveraging DNNs and other AI techniques, smart cities worldwide can better detect and respond to violence, safeguarding lives and properties. The benefits of such technologies are clear; for instance, integrated surveillance systems equipped with advanced AI algorithms can analyze real-time video feeds from CCTV cameras to identify and alert authorities to potential violent incidents, enabling swift intervention. Furthermore, AI-powered predictive analytics can analyze various data sources, including social media feeds and sensor data, to identify patterns and trends associated with violence, enabling authorities to allocate resources strategically and prevent outbreaks of violence in specific areas. As these technologies evolve, they will play an increasingly important role in maintaining safety and security in our cities.  

Current violence detection methods predominantly rely on spatiotemporal models to identify instances of violent activities within video footage. However, it is essential to recognize that violence encompasses a wide range of behaviors, spanning from physical altercations to gunfights. Treating all violent events equally may not effectively prioritize the severity or potential harm involved. To address this challenge, it becomes crucial to develop methods to detect weapons in surveillance footage. 

Existing research has primarily concentrated on identifying weapons using object detection models. Still, these approaches rely on datasets limited to specific weapon types, such as knives and guns. While this provides valuable insights into the detection of known weapons, it fails to account for the reality that virtually any object can be utilized as a weapon in an act of violence. Therefore, there is a pressing need for further research and advancements in open-world weapons detection, which can efficiently identify and classify a broader range of objects that may be employed as weapons. Expanding weapons detection beyond predefined categories equips surveillance systems to recognize potential threats and intervene effectively, improving public safety. Exploring novel approaches and advancements in deep learning, computer vision, and object recognition enables comprehensive open-world weapons detection. These advancements aid in crime prevention and enhance overall community security and well-being.

Furthermore, 3D and spatio-temporal models such as C3D \cite{3.c3d}, I3D \cite{31.i3d} and ConvLSTM \cite{29.Convlstm}, as well as object detection models like You-Only-Look-Once (YOLO) \cite{4.yolo} and RCNN \cite{5.rcnn}, require a high computational load to achieve state-of-the-art performance in violence and weapons detection. This leads to increased carbon footprints in smart cities. We aim to address these challenges by developing a more efficient image classifier capable of accurately detecting weaponizing violence while promoting sustainability in smart cities. To achieve this, we contributed the following:
\begin{itemize}
    \item We address the challenges of detecting violence and weapons in CCTV footage by introducing the \emph{Smart-City CCTV Violence Detection (SCVD)} dataset. Our dataset is designed to facilitate the learning of weapon distribution in surveillance videos, enabling DNNs to effectively detect both weaponized and non-weaponized violence.
    \item We propose \emph{SSIVD-Net} (\textbf{S}alient-Super-\textbf{I}mage for \textbf{Vi}olence \textbf{Det}ection) as a data-centric approach to address the challenges associated with 3D surveillance video in violence recognition tasks. Our approach involves transforming the 3D video data into a Salient-Super-Image representation, reducing data complexity and dimensionality. This transformation enables faster inference, improved performance, and simplified explainability. In particular, our approach allows for seamless integration with 2D vision classifiers, which are not commonly used in the field.
    \item The authors introduce a novel architecture called \emph{Salient-Classifier}, that leverages a kernelized approach with residual networks \cite{6.resnet}. We evaluate variations of SSIVD-Net and Salient-Classifier on our dataset and benchmark against SOTA models used in violence detection. Additionally, we perform comparative analyses to demonstrate the effectiveness of our model using other violence datasets.
\end{itemize}

Through our contributions, we aim to advance the SOTA in violence and weaponized violence detection while also providing a practical and scalable solution for real-world applications.

The rest of the paper is structured as follows: Section 2 discusses the relevant literature on violence detection, Section 3 introduces the SCVD dataset, Section 4 describes the proposed methodology and its components, Section 5 presents the experimental results and comparative analysis, and Section 6 concludes the paper with potential future directions.

\section{Related Work}
\subsection{Weapon Detection}
There are two main approaches to Object Detection: YOLO \cite{4.yolo}, and RCNN \cite{5.rcnn}. YOLO involves taking sliding windows of fixed sizes from the input image at every possible location and feeding them into an image classifier for inference. At the same time, RCNN proposes regions to feed into the classifier. Since their inception, research has gone into optimizing these methods to achieve better performance (\cite{8.mask}, \cite{9.yolov3}), faster inference (\cite{11.fasterrcnn}, \cite{10.yolov2}), or both (\cite{12.fastrcnn}, \cite{13.yolov4}, \cite{14.yolov7}).

To detect weapons in surveillance footage, researchers have leveraged the Faster RCNN object detection model for its high accuracy in identifying objects of interest \cite{15.gunfasterrcnn}, \cite{16.weapon}. However, transfer learning has also been applied to pre-trained Faster RCNN models for detecting handheld guns in clustered scenes \cite{15.gunfasterrcnn}. To classify mostly guns and knives, an ensemble method combining Faster RCNN and Single Shot Detector has been explored \cite{16.weapon}. Another study used a pre-trained YOLO-V4 model on similar datasets \cite{17.bhatti}. However, these methods are limited in their ability to generalize to a wider range of objects that could be used as weapons, and they have not been trained on datasets that include CCTV footage, which limits their ability to detect weapons in various types of video.

\subsection{Violence Detection}
While there are multiple attempts to detect weapons from videos only, there are works that aim to detect violence from videos and CCTV footage. For example, \cite{18.mumtaz} used InceptionNet to detect violence in every frame from sports videos and movies. This results in slower inference and poor generalization results as their method failed to learn temporal properties connecting frames within similar video embeddings. For their models to not lose temporal information, \cite{19.soliman} and \cite{20.Sharma} used ConvLSTMs to detect violence in CCTVs. In ConvLSTMs, an image classifier is employed for spatial feature extraction, while LSTMs learn the temporal information. The authors in \cite{20.Sharma} used a pre-trained ResNet50 model to extract spatial features from the video frames while \cite{19.soliman} leveraged the VGG16 architecture. The extracted features are then concatenated with the latent features from the LSTM \cite{21.lstm}. The above techniques and models require a lot of computing resources to get the results.

Our approach differs from previous techniques as we aim to create a novel data-centric approach that enables image classifiers to learn spatial and temporal features for efficient weaponized violence detection in CCTV videos. We build upon the super image approach, first proposed in \cite{1.superimage}, and showcase Salient-Super-Image to minimize the information lost due to rearranging and resizing video frames into 2D images. Our study is the first to investigate open-world weaponized violence detection in surveillance systems.

\subsection{Super-Image}
The field of action recognition for video classification has been gaining attention in recent years as it plays a significant role in video understanding. Most approaches in this domain use 3D convolutions to classify videos based on appearance, depth, or body skeletons. However, \cite{1.superimage} proposed a different approach using a 2D image classifier (SIFAR). They argued that instead of using deep 3D networks for video action recognition tasks, a simple image classifier could work. To accomplish this, they introduced a technique that involves extracting frames from videos, resizing them, and combining them into a composite image. This super-image effectively captures both local spatial information and global temporal dependencies. The experiments of SIFAR showed promising results, demonstrating that 2D image classifiers could achieve comparable results to their 3D counterparts.

In this work, we take this idea to the next level and unleash \emph{Salient-Super-Image}, an innovative and highly effective variant of the Super Image. We aim to revolutionize the detection of weaponized violence in surveillance systems by preserving crucial information in a simpler yet more impactful manner. Since there are no datasets currently used for this task, we created \textbf{SCVD}, a novel dataset that contains distinctive videos of weaponized, non-weaponized, and normal violence in CCTVs.

\begin{table*}[!h]
\centering
\caption{Comparisons between the SCVD and the previous datasets}
\label{tab:my-table}
\resizebox{\linewidth}{!}{%
\begin{tabular}{@{}l|cccccccc@{}}
\toprule
Dataset & Type & Size & Length/video (sec) & Annotation & Violence & Weapons & Characteristics & Scenario \\ \midrule
NTU CCTV-Fights\cite{22.ntucctv} & Video & 1000 videos & 5-720 & Frame & \textcolor{green}{\cmark} & \textcolor{red}{\xmark}  & CCTV + Mobile   & Natural        \\
Hockey Fight\cite{23.hockey}    & Video & 1000 videos & 1-2   & Video & \textcolor{green}{\cmark} & \textcolor{red}{\xmark}  & Aerial Camera   & Hockey Games   \\
RLVS\cite{19.soliman}             & Video & 2000 videos & 5-15  & Video & \textcolor{green}{\cmark} & \textcolor{red}{\xmark}  & CCTV + Mobile   & Natural        \\
RWF-2000\cite{24.rwf}        & Video & 2000 videos & 5     & Video & \textcolor{green}{\cmark} & \textcolor{red}{\xmark}  & CCTV            & Surveillance   \\
Sohas\cite{25.sohas}           & Image & 3255 images & N/A   & Image & \textcolor{red}{\xmark}  & \textcolor{green}{\cmark} & Captured Images & Demonstrations \\
WVD\cite{26.wvd}             & Video & 168 videos  & 10-72 & Video & \textcolor{red}{\xmark}  & \textcolor{green}{\cmark} & Synthetic       & Computer Games \\
\textbf{SCVD - Ours}            & Video & 500 videos  & 5-10  & Video & \textcolor{green}{\cmark} & \textcolor{green}{\cmark} & CCTV            & Surveillance   \\ \bottomrule
\end{tabular}
}
\end{table*}

\section{Proposed Smart-City CCTV Violence Detection (SCVD) Dataset}

The proposed Smart-City CCTV Violence Detection (SCVD) Dataset is a valuable addition to the existing benchmarks for violence detection, as it is focused explicitly on weaponized violence scenarios captured strictly by surveillance cameras.

The motivation for creating this dataset is to enable AI models to learn, identify, and prioritize the degree of danger in chaotic events, particularly those involving weapons, and alert the authorities accordingly. By identifying weapons in violent events, the authorities can respond more quickly and effectively to prevent further damage and harm. This dataset can also help advance research in the field of AI for violence detection, particularly in the development of models that can effectively identify weapons and distinguish between violent and non-violent events.

Table \ref{tab:my-table} provides a summary of the comparison of the proposed Smart-City CCTV Violence Detection (SCVD) dataset with other datasets used to train deep neural networks (DNNs) for detecting violence and weapons in videos. The datasets considered in this comparison include those used for violence detection, such as Nanyang Technological University (NTU) CCTV-Fights \cite{22.ntucctv}, Hockey Fight \cite{23.hockey}, Real-Life Violent Situations (RLVS) \cite{19.soliman}, and RWF-2000 \cite{24.rwf}, as well as those for weapons detection, including Sohas \cite{25.sohas} and Weapon Violence dataset (WVD) \cite{26.wvd}.

The datasets used for violence detection contain 1000 to 2000 videos, which are annotated either at the frame \cite{22.ntucctv} or video level \cite{23.hockey,19.soliman,24.rwf}. However, these datasets are not properly collected from surveillance systems. Only RWF-2000 \cite{24.rwf} have real-time surveillance videos captured by CCTV surveillance systems. Still, the dataset only contains generalized violence without a distinctive reference to the type that contains weapons, making it also unsuitable for our tasks, which are addressed in this paper.

To detect weapons in videos, Sohas \cite{25.sohas} and Weapon Violence Dataset (WVD) \cite{26.wvd} were created. Sohas contains images with bounding box annotations, while WVD contains synthetic videos of length 10 to 72 seconds generated from the GTA-V computer game and annotated at the video level. Both datasets contain guns, knives, and other well-known weapons, but their distribution is not focused on surveillance. Both datasets focus on only a limited number of weapons as compared to our proposed dataset in this work.

In contrast, our proposed SCVD dataset contains distinctive videos of weaponized and non-weaponized violence scenarios captured from CCTV surveillance systems. In order to enhance the efficiency of weapon recognition, we have opted not to assign specific labels but instead enable DNN models to statistically comprehend the differentiation between scenes involving weapons and those without. Our objective is to assist AI models in gauging the threat level in chaotic events captured by surveillance systems and identifying potential weapons in violent incidents, thereby expediting the intervention of appropriate authorities to prevent further devastation. The SCVD dataset is, therefore, a suitable benchmark for developing and evaluating AI models for weaponized and non-weaponized violence detection in CCTV footage.

Our dataset was obtained by running an automated script that crawled YouTube and downloaded related videos using geographical prompts to ensure that the dataset is not limited to a single location but instead generalized to most parts of the world. The videos were recorded both indoors and outdoors and from various CCTV sensors. The downloaded videos were in 720p resolution with frame dimensions of $1280\times720$. After cleaning the dataset by avoiding noisy and staged videos, the remaining videos' quantity is around 500. These videos were trimmed to 5-10 seconds and annotated at the video level. The final dataset contains three classes: Violence (V), Normal (N), and Weaponized-Violence (WV), with the latter containing videos with an arbitrary number of possible weapons to improve DNNs' performance on weaponized violence detection.

\section{Methodology}
Traditional Convolutional Neural Networks (CNNs) architectures are designed to process individual images, with the object of interest usually located at the center of the image \cite{5.rcnn}, \cite{1.superimage}. However, these approaches are unsuitable for video input as each frame in a video is related to the preceding and succeeding frames. To account for the temporal information in videos, researchers have proposed different methods. One of the approaches is to process each frame individually, neglecting the interdependence between frames, which leads to temporal information loss. Another approach is to extract features from each frame and then use an LSTM networkc\cite{29.Convlstm}, \cite{30.sepconv} to learn the temporal dependencies between these features. However, this method can be computationally expensive and slow.

A more efficient and effective approach is to use 3D Convolutional Neural Networks (3D CNNs), which can directly process spatio-temporal information from videos \cite{3.c3d}, \cite{31.i3d}, \cite{24.rwf}. In 3D CNNs, the filters used for convolution are applied both spatially and temporally, allowing the network to learn features from the entire video sequence. This approach has been successfully applied in various video understanding tasks such as action recognition and video captioning.

\begin{figure*}[t]
    \centering
    \includegraphics[width=\linewidth]{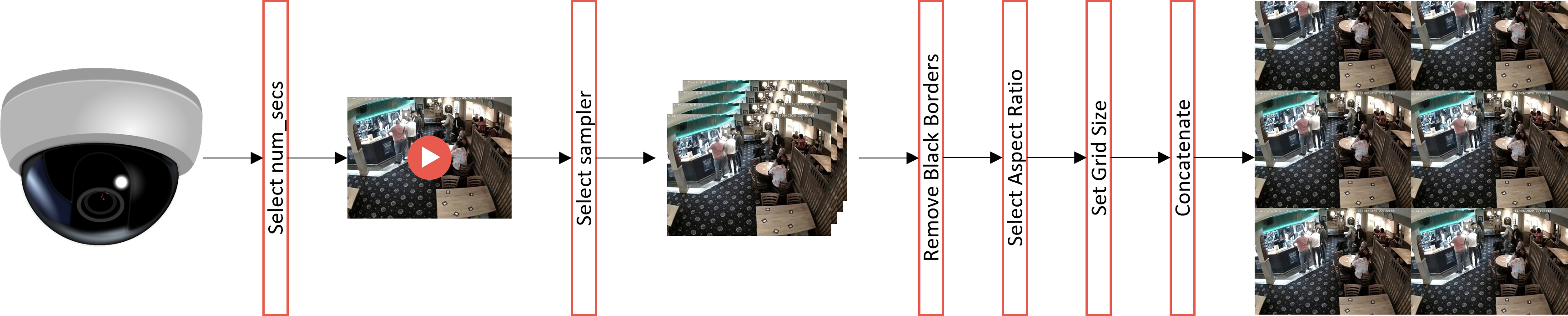}
    \caption{Salient-Super-Image: A sequence of video frames from a CCTV surveillance system are rearranged into a salient-super-image based on given factors such as sample size, sampler, aspect ratio, and spatial arrangement (grid shape).}
    \label{fig:siv}
\end{figure*}

\subsection{Salient-Super-Image}
\textbf{Frame Arrangements}: With video understanding tasks in mind, \cite{1.superimage} explored the possibility of merging multiple input frames from a video to a superimage that contains both spatial and temporal information before processing by CNN. They found that a square formation, i.e., a $4\times4$ arrangement, had the best performance compared to linear or rectangular formations, i.e., a $16\times1$ arrangement. To achieve this, the frames were resized to $224\times224$, combined to form a super image, and then finally resized again to obtain a $224\times224$  image for input into their Swin-Transformer-based SIFAR classifier. However, the super image approach presents certain limitations, particularly in dealing with the original aspect ratio of surveillance videos. Since most videos have a wide/non-square aspect ratio, resizing each frame with dimensions such as $1280\times720$ to a $224\times224$ can introduce noise, result in loss of aspect information by converting the wide frame into a square shape, and potentially hinder performance.

To address this issue, we propose a novel approach called \emph{Salient-Super-Image}. The main idea of our approach is to maintain as much aspect information as possible and reduce information loss in super images by selecting an optimal sample size, sampler, aspect ratio, and spatial arrangement (grid shape) that rearranges the frames while maintaining the aspect ratio of the original frames, i.e., a $1280\times720$, which have rectangular dimensions ($h \times w$, where $w>h$). We aim to achieve a final salient image with $h$ as close to $w$. In mathematical terms, we aim to minimize the difference between the width of $n$ frames and the height of $m$ frames:

\begin{equation}
    \min(w_n-h_m)
\end{equation}
So that:
\begin{equation}
    w_n \approx h_m
\end{equation}

By combining frames using this optimal grid selection, we reduced the information loss. We obtained a $224\times224$ salient image for input into the classifier while still preserving the temporal context of the video (in Figure \ref{fig:siv}, the six frames are uniformly sampled. The $144p\_B$ aspect ratio ($192 \times 144$) was used to resize the images before arranging them in a $3 \times 2$ grid shape).\\

\noindent \textbf{Optimal Frame Selection and Salient Arrangement}: In order to process the videos from our SCVD dataset, which have frame dimensions of $1280 \times 720$, we needed to find an optimal way to merge multiple frames into a single image for input into our \emph{Salient-Classifier}. To achieve this, we considered several factors: 

\begin{enumerate}
    \item \textbf{Sample Size}: The sample size is an essential factor in determining the final grid size for arranging the processed frames. It refers to the number of frames that would like to extract per second, dependent on the selected sampler, and denoted by $k$. It is worth noting that the value of $k$ should be within the range of $3$ and $len(fpgs)$, where $fpgs$ is the total number of frames that can be extracted from a given second(s).
    
    \item \textbf{Sampler}: The type of sampler utilized is also a crucial factor in determining the classifier's performance. To this end, we developed seven types of samplers inspired by existing literature, each designed to select a fixed number of $k$ sample sizes from a given array of frames.

    \begin{itemize}
        \item \textbf{Uniform}: This sampler selects frames uniformly from the video, where the number of frames to select is specified by the parameter $k$. It calculates the stride between the selected frames and then chooses $k$ equally spaced indices according to (Algorithm \ref{uniform}).
        \begin{algorithm}[!h]
        \footnotesize
        \SetAlgoLined
            $stride$ $\gets$ $num\_frames$ // $k$\;
            $indices$ $\gets$ evenly spaced integers from [$0, num\_frames - 1, stride$]\;
            $sampled\_frames$ $\gets$ []]\;
            \For{index in indices}{
                $sampled\_frames.append(frames[index])$\;
            }
            \caption{Uniform Sampler}
            \label{uniform}
        \end{algorithm}
        
        \item \textbf{Random}: This sampler randomly selects $k$ frames from the video without replacement. It uses the function named np.random.choice() to generate a list of $k$ with unique random indices, which correspond to the selected frames according to (Algorithm \ref{random}).
        \begin{algorithm}[!h]
        \footnotesize
        \SetAlgoLined
            $num\_frames$ $\gets$ $len(frames)$\;
            $indices$ $\gets$ randomly select $k$ indices from $[0, num\_frames - 1]$ without replacement\;
            $sampled\_frames$ $\gets$ $[frames[i] \text{ for } i \text{ in } indices]$\;
            \caption{Random Sampler}
            \label{random}
        \end{algorithm}
        
        \item \textbf{Continuous}:This sampler selects $k$ frames that are evenly spaced across the entire video. Initially calculates the stride between the selected frames and then chooses $k$ indices to be evenly spaced across the entire video according to (Algorithm \ref{continuous}).
        \begin{algorithm}[!h]
        \footnotesize
        \SetAlgoLined
            $stride \leftarrow \frac{num\_frames-1}{k-1}$\;
            $indices = []$\;
            \For{$i \leftarrow 0$ \KwTo $k-1$}{ 
                $index_i \leftarrow \lfloor i \times stride \rfloor$\;
                $indices.append[index_i]$}
            $sampled\_frames \leftarrow [frames[i] \text{ for } i \text{ in } indices]$\;
            \caption{Continuous Sampler}
            \label{continuous}
        \end{algorithm}
        
        \item \textbf{Mean absolute difference}: This sampler selects $k$ frames with the smallest average absolute difference between adjacent frames. It calculates the absolute differences between adjacent frames and then selects the $k$ frames with the smallest average absolute difference and returns them in a list according to (Algorithm \ref{mean}).
        \begin{algorithm}[!h]
        \footnotesize
        \SetAlgoLined
            $diffs$ $\leftarrow$ $abs(diff(frames))$\;
            $avg\_diffs$ $\leftarrow$ $mean(diffs)$\;
            $indices$ $\leftarrow$ $argsort(avg\_diffs)[:k]$\;
            $sampled\_frames$ $\leftarrow$ $[frames[i] \text{ for } i \text{ in } indices]$\;
            \caption{Mean Absolute Difference Sampler}
            \label{mean}
        \end{algorithm}
        
        \item \textbf{Lucas-Kanade}: This sampler uses the Lucas-Kanade algorithm to compute optical flow between adjacent frames. It then selects $k$ frames with the largest amount of motion, computes optical flow between adjacent frames, and then selects the $k$ frames with the largest amount of motion according to (Algorithm \ref{lk}).
        \begin{algorithm}[!h]
        \footnotesize
        \SetAlgoLined
            \tcp{Compute optical flow using Lucas-Kanade algorithm}
            $gray\_frames$ $\leftarrow$ [$gray(frame)$ for $frame$ in $frames$]\;
            $sampled\_frames$ $\leftarrow$ $[frames[0]]$\;
            $prev\_frame$ $\leftarrow$ $gray\_frames[0]$\;
            \For{$i \leftarrow 1$ \KwTo $k - 1$}{
                $next\_frame$ $\leftarrow$ $gray\_frames$$[int((i / k) \times (num\_frames - 1))]$\;
                $flow$ $\leftarrow$ $opt\_flow\_farneback(prev\_frame, next\_frame)$\;
                $mag, ang$ $\leftarrow$ $cartToPolar(flow[...,0], flow[...,1])$\;
                $mag$ $\leftarrow$ $normalize(mag)$\;
                $sampled\_frames.append(frames[int((i / k) * (num\_frames - 1))])$\;
                $prev\_frame$ $\leftarrow$ $next\_frame$}
            \caption{Lucas-Kanade Sampler}
            \label{lk}
        \end{algorithm}
        
        \item \textbf{Centered}: This sampler selects $k$ frames centered around the middle of the video. It first selects the middle frame of the video and then selects $k/2$ frames from the first half and $k/2$ frames from the second half. The selected frames are then returned in a list according to (Algorithm \ref{center}).
        \begin{algorithm}[!h]
        \footnotesize
        \SetAlgoLined
            $mid \gets num\_frames$ $//$ $2$\\
            $half\_k \gets k$ $//$ $2$\\
            $stride \gets mid$ $//$ $half\_k$\\
            $indices \gets$ $[i$ $*$ $stride$ $|$ $i$ $\in$ $[0,$ $half\_k)$ $]$\\
            $sampled\_frames$ $+=$ $[frames[i]$ $|$ $i$ $\in$ $indices]$\\
            $indices \gets$ $[i$ $*$ $stride$ $+$ $mid$ $|$ $i$ $\in$ $[0,$ $half\_k)$ $]$\\
            $sampled\_frames$ $+=$ $[frames[i]$ $|$ $i$ $\in$ $indices]$\\
            \caption{Centered Sampler}
            \label{center}
        \end{algorithm}

        \begin{algorithm}[!h]
        \footnotesize
        \SetAlgoLined
            $num\_frames \gets len(frames)$\;
            $sampled\_frames \gets []$\;
            $start \gets random.randint(0, num_frames - k)$\;
            \For{i $\gets$ start \KwTo start+k-1}{
                $sampled\_frames$.append(frames[i]);
            }
            \caption{Consecutive Sampler}
            \label{consecutive}
        \end{algorithm}
        
        \item \textbf{Consecutive}: The consecutive sampler selects a fixed number of consecutive frames from the video frames. For example, suppose we want to select $k$ consecutive frames from a video with $num\_frame$s total frames. In that case, we can start at frame index $i$ and select $k$ frames from that index according to (Algorithm \ref{consecutive}). 
    \end{itemize}

    \item \textbf{Aspect Ratio}: The $aspect\_ratio$ parameter is a tuple of integers that specifies the desired size (width, height) of the cropped and resized frames. These aspect ratios correspond to commonly used video resolutions and dimensions and can be used to specify the desired output size for video frames in a standardized way. The available options for $aspect\_ratio$ include:
    \begin{itemize}
        \item $144p\_A$: $192\times144$ pixels and $240p\_A$: $320\times240$ pixels
        \item $360p\_A$: $480\times360$ pixels and $480p\_A$: $640\times480$ pixels
        \item $144p\_B$: $256\times144$ pixels and $240p\_B$: $426\times240$ pixels
        \item $360p\_B$: $640\times360$ pixels and $480p\_B$: $852\times480$ pixels
        \item $square$: $360\times360$ pixels and $vertical$: $270\times450$ pixels
    \end{itemize}

    \item \textbf{Grid Shape}: Optimizing salient performance also involves integrating an essential factor, namely a tuple of integers whose multiplication should be equivalent to the sample size $k$. When selecting an optimal grid shape, it is crucial to determine suitable row and column coefficients $(r, c)$ based on the width $w$ and height $h$ of the salient-super-image. Ideally, $w \times c$ should be approximately equal to $h \times r$. In the case of our dataset comprising surveillance videos with a $1280\times720$ aspect ratio, the choice of $r$ and $c$ depends on the frame's height $h$ and width $w$ according to the following guideline:
    
    \begin{equation}
    \mathbb{O}_{mn} =
    \begin{cases}
        r > c, \quad \text{if } w > h \\
        r < c, \quad \text{otherwise}
    \end{cases}
    \end{equation}
    This ensures that the optimal selection of $r$ and $c$ aligns with the aspect ratio of the frames.         
\end{enumerate}

\subsection{Salient-Classifier}
We developed a novel deep convolutional neural network called Salient-Classifier to learn the embeddings from the proposed Salient-Super-Images. Salient-Classifier draws inspiration from the ResNet architecture \cite{6.resnet} and Kervolution \cite{28.kerv}. Specifically, the Salient-Classifier takes an input image and passes the signal tensors through a residual-based network architecture similar to ResNet but with several key differences. The input layer, KConv2D, is based on Kervolution's kernel-based convolutional layer. Additionally, we proposed SaliNet, which adds a minimal block and could serve as a sustainable and energy-saving alternative to ResNet's basic and bottleneck blocks.\\

\begin{figure*}[t]
    \centering
    \includegraphics[width=\linewidth]{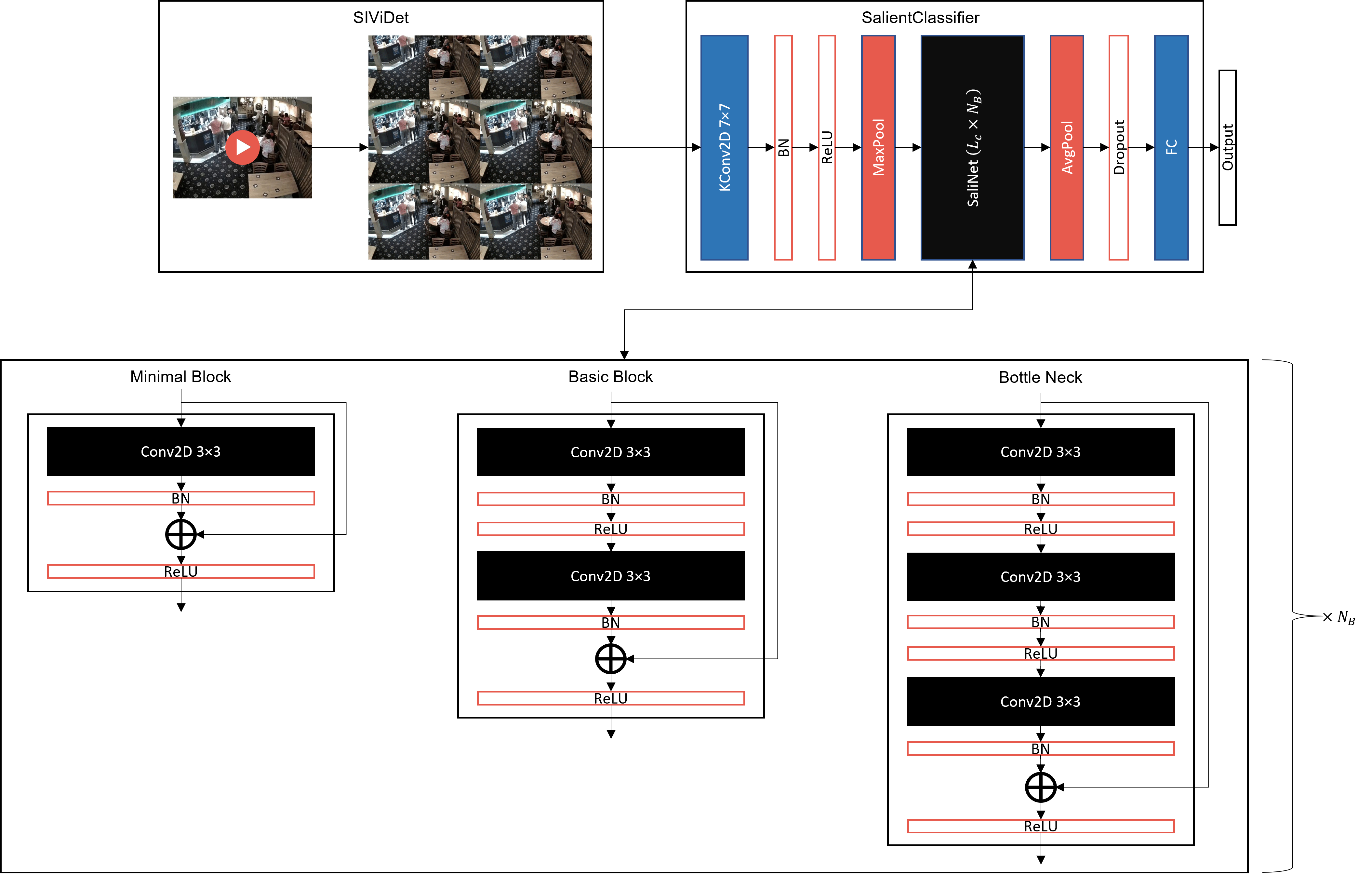}
    \caption{Salient Classifier: The Salient-Super-Images produced in Figure \ref{fig:siv} are fed into our sustainable Salient-Classifier, which follows the residual learning strategy.}
    \label{fig:salient}
\end{figure*}

\begin{figure}[!h]
    \centering
    \includegraphics[height=2.5in]{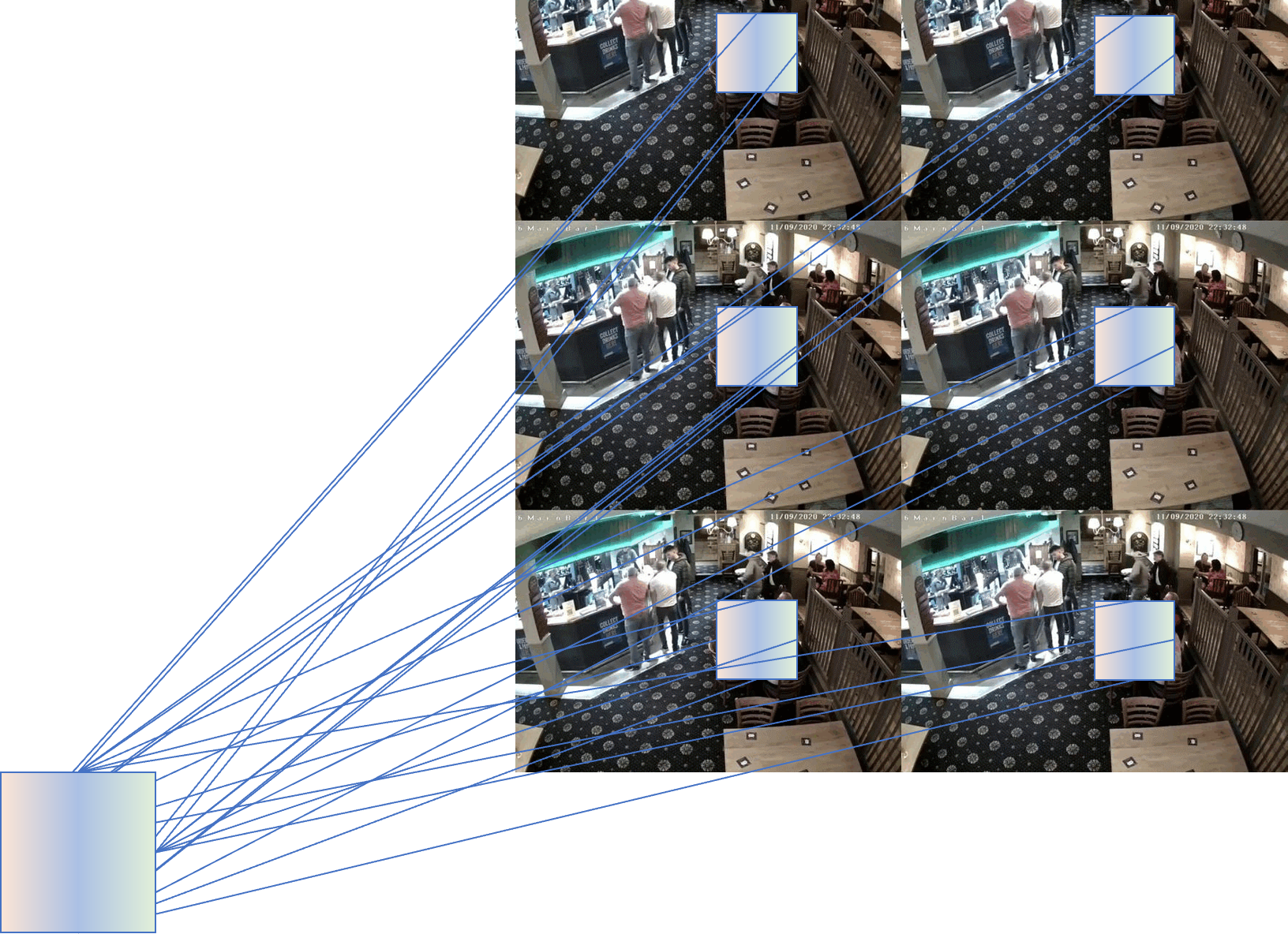}
    \caption{Our model tries to learn similar spatial patterns in the concatenated salient-super-image to avoid the temporal information lost.}
    \label{fig:dyn}
\end{figure}

\noindent \textbf{KConv2D}: In Figure \ref{fig:salient}, the Salient-Super-Image output from SSIVD-Net is fed into a Kernelized Convolution layer (\emph{KConv2D}). As demonstrated in \cite{28.kerv}, the \emph{KConv2D} can effectively learn discriminative features from a larger latent space obtained by expanding the linear outputs $\textbf{x}^T\textbf{w}$ of the vanilla convolution using a given kernel type. However, the proposed SCVD dataset faces a challenge due to the high similarity between the distributions of the violence and weaponized violence classes. This similarity makes it difficult for any Deep Neural Network (DNN) to learn from the dataset. To address this issue, we take inspiration from Support Vector Machines (SVMs), which use kernels such as Sigmoid, RBF, Polynomial, etc., to find a hyperplane in a high-dimensional space that separates data points of different classes with a maximum margin. Thus, this study utilizes two kernels to replace the original convolution kernel to find the maximum distance between these two classes in our dataset. These kernels are:
 \begin{itemize}
     \item \textbf{Polynomial Kernel}: In Support Vector Machines (SVMs), the polynomial kernel is widely used for non-linear classification. It maps data points to a higher-dimensional feature space using a polynomial degree that controls the complexity of the decision boundary. In this work, the authors employ a polynomial kernel instead of a linear kernel in convolution layers. Specifically, we compute the polynomial kernel as:
     \begin{equation}
        K(x, w) = (x^Tw + c)^d
    \end{equation}
    where $c$ is a learnable balance factor, and $d$ is the degree of the polynomial. Notably, we make $c$ learnable instead of a constant in the original polynomial kernel to control the influence of individual data points and help with an expressive translation to improve and shift the decision boundary between close classes. The kernel function takes two inputs, $x$ and $w$, representing data points in the input space and maps them to a higher-dimensional feature space. In this feature space, the polynomial kernel computes the dot product of the mapped features. The degree $d$ determines the polynomial and controls the complexity of the decision boundary. Nevertheless, if the degree is too high, it can lead to overfitting.

    \item \textbf{Gaussian Kernel}: In SVMs, the RBF (Radial Basis Function) kernel maps the input space to a higher dimensional feature space, which is more likely to be linearly separable. The feature space is defined by the distance between the input data points and a set of reference points, also known as support vectors. The RBF kernel function measures the similarity between the input data points and these support vectors based on their distance in this higher dimensional space. The RBF kernel function is defined as:
    \begin{equation}
        K(x, x') = \exp(-\gamma ||x - x'||^2)
    \end{equation}  
    where $x$ and $x'$ are two input data points, $\gamma$ is a parameter that controls the width of the RBF kernel, and $||x - x'||^2$ is the squared Euclidean distance between $x$ and $x'$. The kernel function outputs a similarity score between the two input data points.

    For the proposed network, the Gaussian kernel, which is based on the RBF definition, is written as:
    \begin{equation}
        K(x, w) = \exp(-\gamma ||x - w||^2)
    \end{equation}
    Here, we replace the reference points with the weights.
 \end{itemize}

\noindent \textbf{SaliNet}: We propose the \emph{\textbf{M}inimal-Block (m)}, while retaining the \emph{\textbf{B}asic-Block (b)} and \emph{Bottle-\textbf{N}eck (n)} of ResNets to make up our Salient Block (\emph{SaliNet}). While the BottleNeck and Basic Block have three and two convolution layers, respectively, the minimal only requires one, which uses a $3\times3$ filter followed by batch normalization before the residual comes in. The output of these computations is finally activated using the ReLU function. Also, we proposed new layer arrangements for each adopted block type to satisfy our requirement of creating networks requiring less energy and computation. Table \ref{salinet} shows the information about these blocks.

With our design approach, the Salient-Classifier harnesses the power of KConv2D to extract similar spatial patterns within each Salient-Super-Image (see Figure \ref{fig:dyn}). These learned dynamics are subsequently fed into the SaliNet blocks, enabling the model to acquire more valuable features. This design enhances the accuracy of the classifier and facilitates efficient inference performance.

\begin{table}[!h]
\caption{The Salient-Classifier's SaliNet has three block styles: The Basic Block(b) and BottleNeck(n), which were adapted from the ResNet architecture, as well as our novel Minimal Block(m). This table shows the layer arrangement for each SaliNet block style and the number of parameters based on the chosen arrangement and style.}
\label{salinet}
\resizebox{\linewidth}{!}{%
\begin{tabular}{@{}cc|ccc@{}}
\toprule
\multicolumn{1}{l}{} & \multicolumn{1}{l}{} & \multicolumn{1}{l}{} & \multicolumn{1}{l}{Number of Params (M)} & \multicolumn{1}{l}{} \\ 
Classifier           & Layer Arrangement    & Minimal Block(m)     & Basic Block(b)                           & Bottle Neck(n)       \\ \midrule
SaliNet-2            & 1,1,0,0              & 1.8                  & 4.9                                      & 8.0                  \\
SaliNet-4            & 1,1,1,1              & 1.8                  & 4.9                                      & 8.0                  \\
SaliNet-8            & 2, 2, 2, 2           & 4.9                  & 11.2                                     & 14.0                 \\
SaliNet-16           & 3, 4, 6, 3           & 10.0                 & 21.3                                     & 23.5                 \\ \bottomrule
\end{tabular}%
}
\end{table}

\section{Experiments and Results}
\subsection{Environment}
\noindent \textbf{Setup}: The following details show the environment setup for the experiments conducted for this research:

\begin{itemize}
    \item System used: NVIDIA Quadro RTX6000, with GPU Memory: 24GB GDDR6 having Ubuntu 21.04 Operating system.
    \item Libraries: Tensorflow\footnote{https://github.com/tensorflow/tensorflow}, Keras\footnote{https://github.com/keras-team/keras}, Pytorch\footnote{https://github.com/pytorch/pytorch}
\end{itemize}

\noindent \textbf{Training Details}: To train all networks, including baselines on the SCVD dataset, we used the Stochastic Gradient Descent (SGD optimizer) with a learning rate of $1e-3$ and a momentum of $0.9$. The optimizer's choice is based on the general knowledge that SGDM ensures full convergence, compared to Adam. The SGD optimizer for parameter update is:

\begin{equation}
   w_{t+1}=w_{t}-\alpha\frac{\partial L}{\partial w_{t}}
\end{equation}

\noindent where $w_{t}$ denotes the weight, $w_{t+1}$ denotes the updated parameter, $\alpha$ denotes the learning rate, and $\frac{\partial L}{\partial w_{t}}$ denotes the partial derivative of the gradient. To ensure fairness in comparison, all Salient-Classifier variants were initialized with the same weights and trained from scratch for 30 epochs. We recorded the average test accuracy and average precision for each network.

\subsection{Results and Discussions}
\noindent \textbf{Baseline Results}: We conducted a comprehensive hyperparameters search for the Salient-Classifier module using the smallest and fastest block variant, SaliNet-2m. Our iterative approach involved experimenting with all possible hyperparameters to identify the best-performing option. This allowed us to eliminate inferior options and ultimately obtain the optimal hyperparameters for subsequent experiments. Table \ref{baseline} shows the results.

\begin{table}[!h]
\small
\caption{Table showing results of different hyperparameters for the SaliNet-2m. We employed an elimination routine to select the best hyperparameters for the proposed SCVD dataset.}
\label{baseline}
\resizebox{\linewidth}{!}{%
\centering
\footnotesize
\begin{tabular}{@{}cccccc@{}}
\toprule
k - grid\_shape & Sampler     & Aspect Ratio & Accuracy(\%) & AP(\%)   & Inference time (s) \\ \midrule
4 - 2x2         & \textbf{uniform}     & \textbf{square}       & \textbf{78.4}     & \textbf{80.5} & \textbf{0.04}               \\
4 - 2x2         & random      & square       & 75.5     & 79.9 & 0.05               \\
4 - 2x2         & continuous  & square       & 74.4     & 76.2 & 0.04               \\
4 - 2x2         & mean\_abs   & square       & 71.1     & 77.7 & 0.15               \\
4 - 2x2         & LK          & square       & 69.6     & 78.2 & 0.21               \\
4 - 2x2         & centered    & square       & 73.2     & 78.7 & 0.04               \\
4 - 2x2         & consecutive & square       & 70.4     & 79.4 & 0.04               \\ \midrule
6 - 3x2         & uniform     & 144p\_A      & 78.9     & 81.2 & 0.05               \\
6 - 3x2         & uniform     & 144p\_B      & 79.7     & 81.9 & 0.05               \\
6 - 3x2         & uniform     & 240p\_A      & 80.9     & 84.0 & 0.05               \\
6 - 3x2         & uniform     & 240p\_B      & 81.3     & 84.2 & 0.05               \\
6 - 3x2         & uniform     & 360p\_A      & 78.4     & 81.9 & 0.05               \\
6 - 3x2         & uniform     & 360p\_B      & 82.4     & 83.8 & 0.05               \\
6 - 3x2         & uniform     & \textbf{480p\_A}      & \textbf{83.0}     & \textbf{83.4} & \textbf{0.05}               \\
6 - 3x2         & uniform     & 480p\_B      & 83.0     & 83.4 & 0.06               \\ \midrule
9 - 3x3         & uniform     & square       & 84.7     & 85.0 & 0.06               \\
12 - 4x3        & uniform     & 480p\_A      & \textbf{86.6}     & \textbf{89.6} & 0.07               \\
15 - 5x3        & uniform     & 480p\_A      & 84.3     & 86.8 & 0.08               \\ \bottomrule
\end{tabular}%
}
\end{table}

We began by utilizing various samplers to select $4$ frames from the input videos using a square aspect ratio. Among these samplers, the uniform sampler (\ref{uniform}) outperformed the others, achieving an accuracy of 78.4\% and an average precision of 80.5\%, and was thus used in all subsequent experiments. We then proceeded to use $6$ frames with different aspect ratios, with 480p\_A and 480p\_B achieving the best results, achieving an accuracy of 83.0\% and an average precision of 83.4\%. Of these, 480p\_A achieved a slightly faster average inference time per salient-super-image than 480p\_B.

Finally, we used the uniform sampler and 480p\_A aspect ratio to select 12 and 15 frames from the input videos, for which we achieved the best accuracy of 86.6\% and average precision of 89.6\% on 12 frames. Furthermore, our results demonstrate the superiority of our \emph{salient-super-image} approach over the original \emph{superimage} method \cite{1.superimage}. The authors of the \emph{superimage} approach used a consecutive frame sampler with a square aspect ratio. This supports our hypothesis that using \emph{superimage} is limiting when analyzing wide-aspect videos extracted from surveillance systems.\\

\noindent \textbf{Comparison with SOTA approaches on the SCVD dataset}: Table \ref{compare} compares the performances of various models on the SCVD dataset. We can observe that the proposed \emph{Salient-Classifer} variants achieve highly competitive results compared to existing methods.

\begin{table}[!h]
\small
\centering
\caption{Comparisons between current SOTA models in violence detection and variants of Salient-Classifiers on the SCVD dataset.}
\label{compare}
\footnotesize
\begin{tabular}{@{}lcc@{}}
\toprule
Model                           & Num\_Params (M) & Accuracy (\%) \\ \midrule
FGN \cite{24.rwf}               & \textbf{0.3}             & 74.4          \\
Conv-LSTM \cite{29.Convlstm}    & 47.4            & 71.6          \\
Sep-Conv-LSTM \cite{30.sepconv} & 0.4             & 78.4          \\ \midrule
SaliNet-2m                      & 1.8             & \textbf{86.6}          \\
SaliNet-4m                      & 1.8             & 83.1          \\
SaliNet-8m                      & 4.9             & 77.8          \\
SaliNet-2b                      & 4.9             & 75.9          \\
SaliNet-2n                      & 8.0             & 78.8          \\ \bottomrule
\end{tabular}%
\end{table}

Initially, we observe that the Flow-Gated-Network (FGN) model \cite{24.rwf}, which incorporates 3D convolutions, achieved a maximum accuracy of 74.4\%. The LSTM-based networks, including ConvLSTM \cite{29.Convlstm} and SepConvLSTM \cite{30.sepconv} attained an accuracy of 71.6\% and 78.4\%, respectively. These LSTM models utilize dynamic 2D convolution filters from pre-trained models to extract spatial features from each frame, combined to obtain spatio-temporal features for inference. While these models exhibit a relatively lower parameter count, they encounter challenges in capturing the nuanced temporal dynamics that differentiate between weaponized and non-weaponized violence classes.

In contrast to the models mentioned above, our proposed Salient-Classifier variants, which employ the SaliNet module with different block styles, demonstrated superior performance. Notably, the minimal block variants exhibited a greater capacity to capture critical information, enabling effective differentiation between occluding classes. Among our Salient-Classifier models, those utilizing the SaliNet-2m and SaliNet-4m achieved remarkable accuracy of 86.6\% and 83.1\%, respectively, while maintaining compact sizes of only 1.8 million parameters each. It is worth mentioning that the remaining variants displayed performances comparable to the current SOTA models, i.e., Flow-Gated-Net \cite{24.rwf} and Separable-ConvLSTMs \cite{30.sepconv}.

Moreover, we note that increasing the model complexity by utilizing larger block styles, such as SaliNet-8m, SaliNet-2b, and SaliNet-2n, does not necessarily improve the Salient-Classifier's performance. This observation indicates that the optimal trade-off between model capacity and complexity is achieved with SaliNet-2m. However, further investigation is required in future work to understand the underlying reasons for this phenomenon. Nevertheless, these results highlight the effectiveness of our proposed network in capturing pertinent spatial features within intricate classes in our dataset. Additionally, the compact size of the model contributes to its computational efficiency, rendering it suitable for real-world applications.\\

\noindent \textbf{Comparison with SOTA approaches on benchmark datasets}: We conducted a comprehensive performance comparison among various variants of our classifiers, including 2m, 2b, and 2n, alongside well-established 3D CNN architectures such as C3D, I3D, and FGN. Additionally, we evaluated architectures based on CNN-LSTM hybrids, namely Conv-LSTM, Bi-Conv-LSTM, and Sep-Conv-LSTM. The evaluation encompassed three distinct datasets: MovieFight \cite{23.hockey}, HockeyFight \cite{23.hockey}, and RWF-2000 \cite{24.rwf}. The detailed comparative analysis of these techniques with the proposed is mentioned in Table \ref{comp}.

\begin{table}[!h]
\small
\centering
\caption{Comparison between current SOTA models in violence detection and variants of salient classifiers on benchmark datasets}
\label{comp}
\footnotesize
\begin{tabular}{@{}llccc@{}}
\toprule
Method             & Model         & MovieFight & HockeyFight & RWF-2000 \\ \midrule
                   & C3D           & 100.0      & 96.5        & 82.8 \\
3D-CNNs            & I3D           & 100.0      & 98.5        & 85.8 \\
                   & FGN           & 100.0      & 98.0        & 87.3 \\ \midrule
                   & Conv-LSTM     & 100.0      & 97.1        & 77.0 \\
Conv-LSTM          & Bi-Conv-LSTM  & 100.0      & 98.1        & -    \\
                   & Sep-Conv-LSTM & 100.0      & 99.5        & 89.3 \\ \midrule
                   & SaliNet-2m    & 100.0      & \textbf{100.0}       & 88.5 \\ 
Salient-Classifiers & SaliNet-2b    & 100.0      & \textbf{100.0}       & 89.7 \\
                   & SaliNet-2n    & 100.0      & \textbf{100.0}       & \textbf{90.3} \\ \bottomrule
\end{tabular}%
\end{table}

We trained our proposed classifiers on the MovieFight and HockeyFight datasets \cite{23.hockey} using the 480p\_A aspect ratio. Since these datasets do not have a consistent number of frames per second (fps) across all videos, we utilized the uniform sampler to extract 12 fps with a grid shape of $(4,3)$. On the other hand, for the SCVD dataset, all videos have a fixed frame rate of 30 fps. Therefore, we utilized all the frames to create a Salient-Super-Image with a 144p\_A aspect ratio and a grid shape of $(6,5)$. We conducted training for 30 epochs on the MovieFight and HockeyFight datasets and 50 epochs on the RWF-2000 dataset to ensure sufficient model convergence.

Based on the results presented in Table \ref{comp}, our classifiers utilizing the Salient-Super-Image approach outperformed the other approaches across multiple datasets. Notably, our classifiers achieved a perfect accuracy of 100\% on the MovieFight dataset, demonstrating their robustness in capturing the distinguishing characteristics of fight scenes in movies. Additionally, our classifiers achieved a perfect accuracy of 100\% on the HockeyFight dataset, indicating their effectiveness in accurately detecting fights in hockey videos.

On the more challenging RWF-2000 dataset, our best-performing model achieved an impressive accuracy of 90.3\%. This showcases the superior performance of our approach in identifying violence in surveillance videos, where the presence of occlusions, varying lighting conditions, and other factors make the task more complex. These results highlight the efficacy of our Salient-Super-Image approach in enhancing the performance of violence detection classifiers across diverse datasets. The high accuracies achieved by the Salient-Classifier's variations demonstrate the potential of our approach for real-world applications in smart surveillance systems.

\section{Conclusion}
In this study, we aimed to accurately distinguish weaponized violence from non-weaponized ones within surveillance systems while simplifying the architecture involved. To address the scarcity of specialized video datasets for weaponized violence, we introduced the innovative \emph{SCVD} dataset. Our proposed \emph{Salient-Super-Image} technique, which transforms the 3D video analysis task into a 2D perspective for the use of selected image classifiers, significantly improved violence detection efficiency. Our experimental results revealed that the SSI technique effectively mitigates information loss compared to the Super-Image approach \cite{1.superimage}. Additionally, the introduced \emph{Salient-Classifiers} outperformed state-of-the-art methods on the SCVD dataset, and our approaches were further validated on benchmark datasets. Our contributions not only underscore the benefits of our approaches but also underscore the importance of addressing the unique challenges associated with weaponized violence detection, opening the door to more precise and efficient violence detection systems that can enhance public safety and contribute to the advancement of smarter and safer cities.

\bibliographystyle{splncs03_unsrt} 
\bibliography{egbib}

\end{document}